# Eigen-CAM: Class Activation Map using Principal Components

Mohammed Bany Muhammad       Mohammed Yeasin

University of Memphis

{mbnymhmm, myeasin }@memphis.edu

**Abstract—** Deep neural networks are ubiquitous due to the ease of developing models and their influence on other domains. At the heart of this progress is convolutional neural networks (CNNs) that are capable of learning representations or features given a set of data. Making sense of such complex models (i.e., millions of parameters and hundreds of layers) remains challenging for developers as well as the end-users. This is partially due to the lack of tools or interfaces capable of providing interpretability and transparency. A growing body of literature, for example, class activation map (CAM), focuses on making sense of what a model learns from the data or why it behaves poorly in a given task. This paper builds on previous ideas to cope with the increasing demand for interpretable, robust, and transparent models. Our approach provides a simpler and intuitive (or familiar) way of generating CAM. The proposed Eigen-CAM computes and visualizes the principle components of the learned features/representations from the convolutional layers. Empirical studies were performed to compare the Eigen-CAM with the state-of-the-art methods (such as Grad-CAM, Grad-CAM++, CNN-fixations) by evaluating on benchmark datasets such as weakly-supervised localization and localizing objects in the presence of adversarial noise. Eigen-CAM was found to be robust against classification errors made by fully connected layers in CNNs, does not rely on the backpropagation of gradients, class relevance score, maximum activation locations, or any other form of weighting features. In addition, it works with all CNN models without the need to modify layers or retrain models. Empirical results show up to 12% improvement over the best method among the methods compared on weakly supervised object localization.

**Keywords—** Weakly supervised localization, Visual explanation of CNN, Explainable AI, Class activation maps, Salient features.

## I. INTRODUCTION

The trade-off between generalization and interpretability [1], accuracy versus simplicity is a common theme of any machine learning algorithm. Usually, easy to interpret algorithms tend to generalize poorly on unseen data as in the case of Decision trees [2] and K-means [3] algorithms, and hard to interpret algorithms usually generalize better as in the case of most deep learning (DL) models with millions of parameters [4], [5].

Visual explanation of accurate and non-interpretable CNN-based deep models can help the end-users in bridging the gap between model generalization and interpretability. It can identify sources of prediction failures that may help improved performances and building trust in complex DL models.

The widely used convolutional neural network (CNN) based architectures in DL are proven to be the best in learning representations and solving complex computer vision and general artificial intelligence problems such as image classification [4], [5], object localization [6]–[8], semantic segmentation [9]–[12], image captioning [13]–[16]and visual question answering (VQA) [17], [18]. CNN architecture is composed of a cascade of various types of layers, such as convolution, MaxPooling, drop-out, fully connected, and SoftMax. The nonlinearity allows a higher degree of freedom, drawing the decision boundaries to improve generalization and, at the same time, causes a lack of interpretability; in other words, no direct reversible input-output relationship.

Visual explanations of CNN are expected to provide a high-resolution class discriminative interpretation for various tasks. A plethora of methods, such as CAM [19], Grad-CAM [20], Grad-CAM++ [21], CNN-fixations [22] have reported results with varying degrees of success. The overall accuracies of these methods remain low in certain applications such as weakly supervised localization and fail to provide explanations for misclassified examples. In addition, methods like CNN fixations require many operations to memorize important activations in every layer.

The proposed Eigen-CAM uses the principal components of the learned representations from the convolutional layers to create the visual explanations. The major contributions are:

- We present a simple, intuitive method to obtain CAM based on convolutional layers output, and the process is independent of class relevance score.

- We demonstrate that the proposed Eigen-CAM can robustly and reliably localize objects without the need to modify CNN architecture or even to backpropagate any computations, and at the same time, achieves higher performance compared to all previously reported methods such as Grad-CAM, CNN fixations.

Figure 1 shows examples depicting the ability of Eigen-CAM in generating visual explanations for multiple objects in an image. Eigen-CAM was found to be robust against classification errors made by fully connected layers in CNNs, does not rely on the backpropagation of gradients, class relevance score, maximum activation locations, or any other form of weighting features.

The rest of the paper is organized as follows. In Section II, we present related reported literature to provide the research context. Following this, we present the details of the proposed Eigen-CAM in section III. Subsequently, in Section IV, we present the results of the empirical evaluation of Eigen-CAM and compare and contrast the performance against state-of-the-art methods across different applications. Finally, Section V concludes the paper with a few remarks on lessons learned and future directions.

## II. RESEARCH CONTEXT

CNN visualizations utilize a single forward pass to produce learned representation in an effort to interpret CNN models. Such methods can be divided into two groups, namely class non-discriminative and class discriminative representations. The former produce results capable of identifying salience features in the input space, while the latter method identifies parts of the salience features in the input space responsible for a particular response at the output of any DL model. The latter class of visualization explains the output decision by locating the region in the input space that causes this decision.

Class non-discriminative CNN visualization backpropagate gradients from class score (input to SoftMax) to the input space (image pixels), to locate distinct features, for example, activation maximization [23] intended for answering the question of, what input may have maximized the output? Such methods could be used across the whole model or a single layer in the CNN network. The saliency maps [23] answers a very different question. Given a model and an input image, what part of the input image may have maximized the class score? DeconvNet [23], [24] and guided backpropagation [25]attempt to answer the same question addressed by saliency map, and differs from saliency map by the way they handle the nonlinearity resulting from MaxPooling. DeconvNet replaces the MaxPooling layer by using higher values for strides at the convolution layer to reduce the total number of features extracted across the CNN. At the same time, guided backpropagation, which outperforms all previous methods achieved that mainly by adding constraints on what gradients should be backpropagated.

Class discriminative CNN visualization methods can be grouped into two broad categories based-relevance score and gradient propagation. The former propagates the probability of SoftMax layer all the way back to the input space to find the relevance of each input pixel, an excellent example of those methods are Deep LIFT [26] and Layer-Wise Relevance Propagation [27], while gradient-based methods backpropagate Gradients to last convolution layer to obtain class activation map (CAM),

The method CAM was the first method to locate class activation map [19], by removing the fully connected layers at the end of any CNN model and replacing the MaxPooling layer

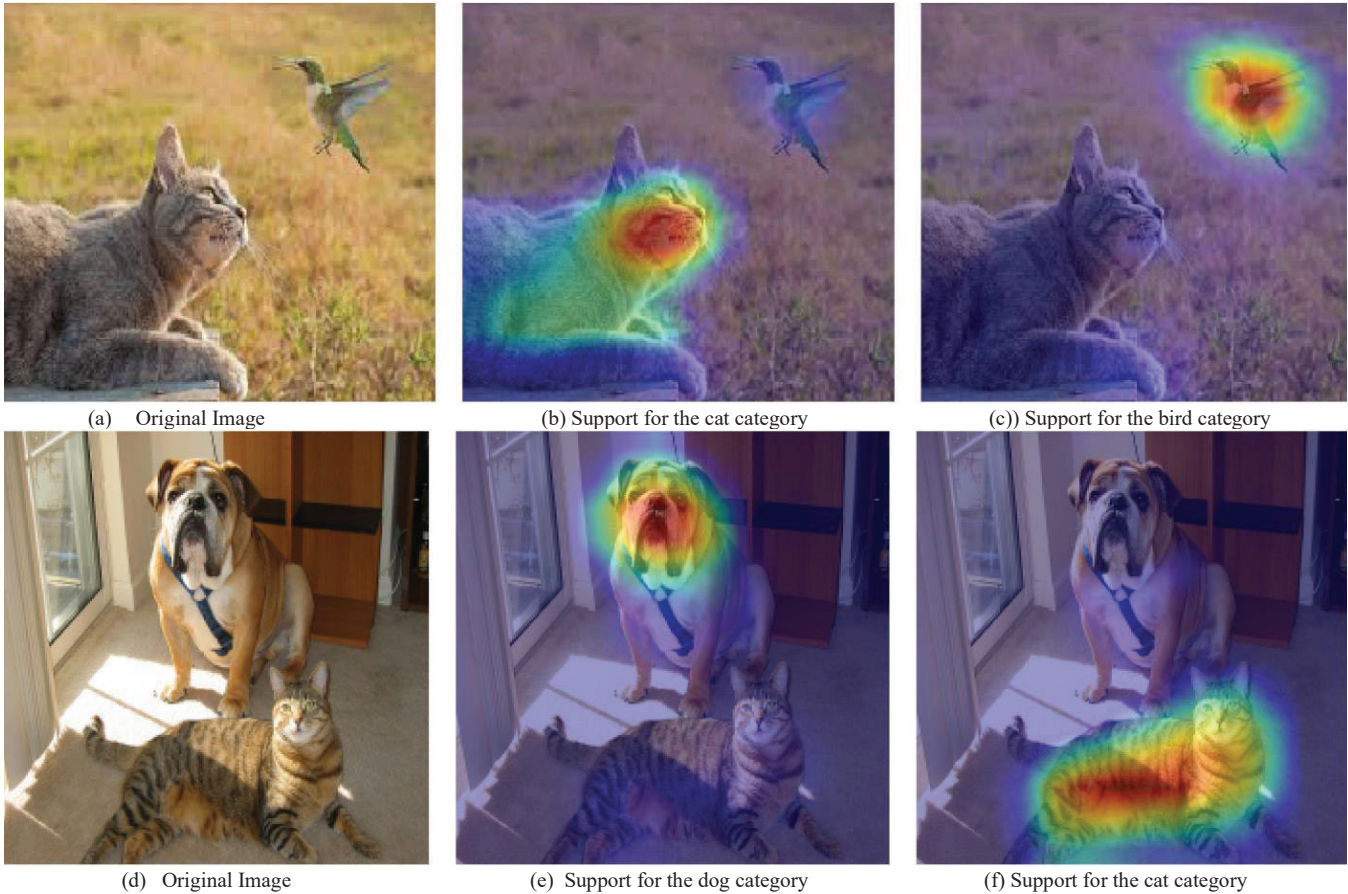

(a) Original Image     (b) Support for the cat category     (c)) Support for the bird category

(d) Original Image     (e) Support for the dog category     (f) Support for the cat category

Fig. 1: Eigen-CAM visualizations computed for two sample images. (a, d), Image. (b, c, e, f) shows class activations computed using Eigen-CAM, middle column represent support for different categories obtained using the fist principle component and right column represent support for different categories obtained using the second principle component.

by a global average pooling (GAP) layer and by calculating the weighted average of features extracted by the feature extraction network a class activation map is obtained.

Even though CAM did not use gradient to weight features produced by the last convolutional layer to obtain the class activation map, it inspires methods like Grad-CAM and Grad-CAM++, where Grad-CAM generalizes CAM to all different CNN models without the need to change or modify the CNN model and relies on gradients to weight features learned at the last convolutional layer. Grad-CAM++ enhances Grad-CAM visualization through pixel-wise weighting of the gradients to enhance multiple class occurrence and refine the process of generating feature maps.

One of the recently reported methods called, CNN Fixations that did not fall under any previous categorizations, is inspired by the biological vision, like the way human eyes perform fixations. CNN Fixations track strong activation from the class space back to image space, providing a high resolution, pixel-level localizations. This idea depends on the accuracy of the predicted class, as the class score increases, the fixation points tend to fall more inside the object triggering that class.

To summarize, all class discriminative methods require propagating gradients in the case of Grad-CAM or Grad-CAM++, score probability in the case of DeepLIFT, and Layer-Wise Relevance Propagation or strong activation in the case, of CNN Fixations. Backpropagating any quantity requires addition computational overhead and assumes that classifiers produced correct decisions, and whenever a wrong decision is made, all mentioned methods will produce wrong or distorted visualization as can be seen in Figure 3.

To address the above-mentioned problems, we present Eigen-CAM that is intuitive and compatible with all DL models without any modifications.

### III. PROPOSED APPROACH

CNN-based deep neural networks outperform all other methods due to the ability to learn spatial relationships between image pixels. In general, CNN architecture can be divided into feature extraction network and classification network.

The main component in the feature extraction network is convolutional layers, where the convolutional layers at the early stages learn lower level spatial features such as edges and corners. Along with the hierarchical structure, convolutional layers learn higher levels of abstractions and possibly features that can produce semantic meaning, at least to the categorical level. On the other hand, the classification network is the fully connected layers that flatten the learned features and draw decision boundary to classify the concepts. Let us start with the following observations to have an intuitive feel and the need of the Eigen-CAM:

Observation 1: All methods that implement backpropagation of quantities such as gradients, relevance score, or maximum activation location, implicitly assume that the CNN model is 100% accurate. But these methods work only when the classification result is correct. It is possible that learned representation is correct, and the classification result is inaccurate due to limitations in learning the non-linear decision surface and vice-versa.

Observation 2: CAM uses the last weight matrix between GAP and SoftMax to weight different feature maps. Similarly, in Grad-CAM and Grad-CAM++ derives the weights of the linear combination of different feature maps, based on backpropagated class relevance score. Radiant values determine the weight of each feature map to produce the class activation map.

Observation 3: We know that gradients are noisy by nature. In addition, Grad-CAM and Grad-CAM++ use only positive gradients to weigh features and assign higher weights for larger gradients regardless of redundancy in the feature space or shape of the manifold in the feature space.

Observation 4: CNN maps images into different classes; the mapping matrix (Model) is expected to learn salient features. During the learning process, optimizers adjust the weights of filters in convolution layers to extract distinct features and adjust the weights of fully connected layers to classify extracted features. Complex hierarchical representation is mapped onto the last convolutional layer, and the decision boundary is learned using fully connected layers.

So, the question now is what features go through all local linear transformations and stay relevant in the same direction of maximum variation. In other words, what salience features will be in the direction of the principal component of the learned representation.

We assume that all relevant spatial features in the input image learned over the hierarchy of the CNN model will be preserved during the optimization process, and non-relevant features will be regularized or smoothed out.

let I represent the input image of size $(i \times j)$ $I \in R^{i,j}$, and let $W_{L=n}$ represent the combined weight matrix of the first k layers of size (m, n).

The class activated output is the image I projected onto the the last convolution layer L=k and is given by

$$O_{L=k} = W_{L=k}^T I \quad (1)$$

Factorizing $O_{L=k}$ using singular value decomposition to compute the principal components of $O_{L=k}$ gives

$$O_{L=k} = U\Sigma V^T \quad (2)$$

Where U is an $M \times M$ orthogonal matrix and the column of U are the left singular vectors, $\Sigma$ is a diagonal matrix of size $M \times N$ with singular values along the diagonal and V is an $N \times N$ orthogonal matrix and the column of V are the left singular vectors.

The class activation map, $L_{Eigen\text{-}CAM}$ is given by the projection of $O_{L=k}$ on the first eigenvector

$$L_{Eigen\text{-}CAM} = O_{L=k} V_1 \quad (3)$$

where $V_1$ is the first eigen vector in the V matrix.

## IV. APPLICATIONS

In this section, we will show that Eigen-CAM is a general visualization method capable of working with any framework or network without the need to modify, train, or backpropagate any parameter across layers. We evaluated Eigen-CAM against state-of-the-art reported methods that include multiple computer vision tasks and modalities across different applications.

In Figure 2 we demonstrate the Eigen-CAM capability in detecting class activation maps (CAM), Figure 2 shows better capability of localizing discriminative regions compared to other methods and shows better localization consistency in different scenario's like single and multiple object detection, detecting object in the foreground or the background of an image and finally detecting objects in images with crowded or plain background.

In Figure 3 we presented two misclassified examples from ILSVRC validation set, top-3 classification results for the image in figure 3.a using VGG-16 were (Dogsled with probability 0.543, Collie with probability 0.133 and Groenendael with probability 0.088), figure 3.d shows that Eigen-CAM successfully localizes features for a Collie, unlike other methods.

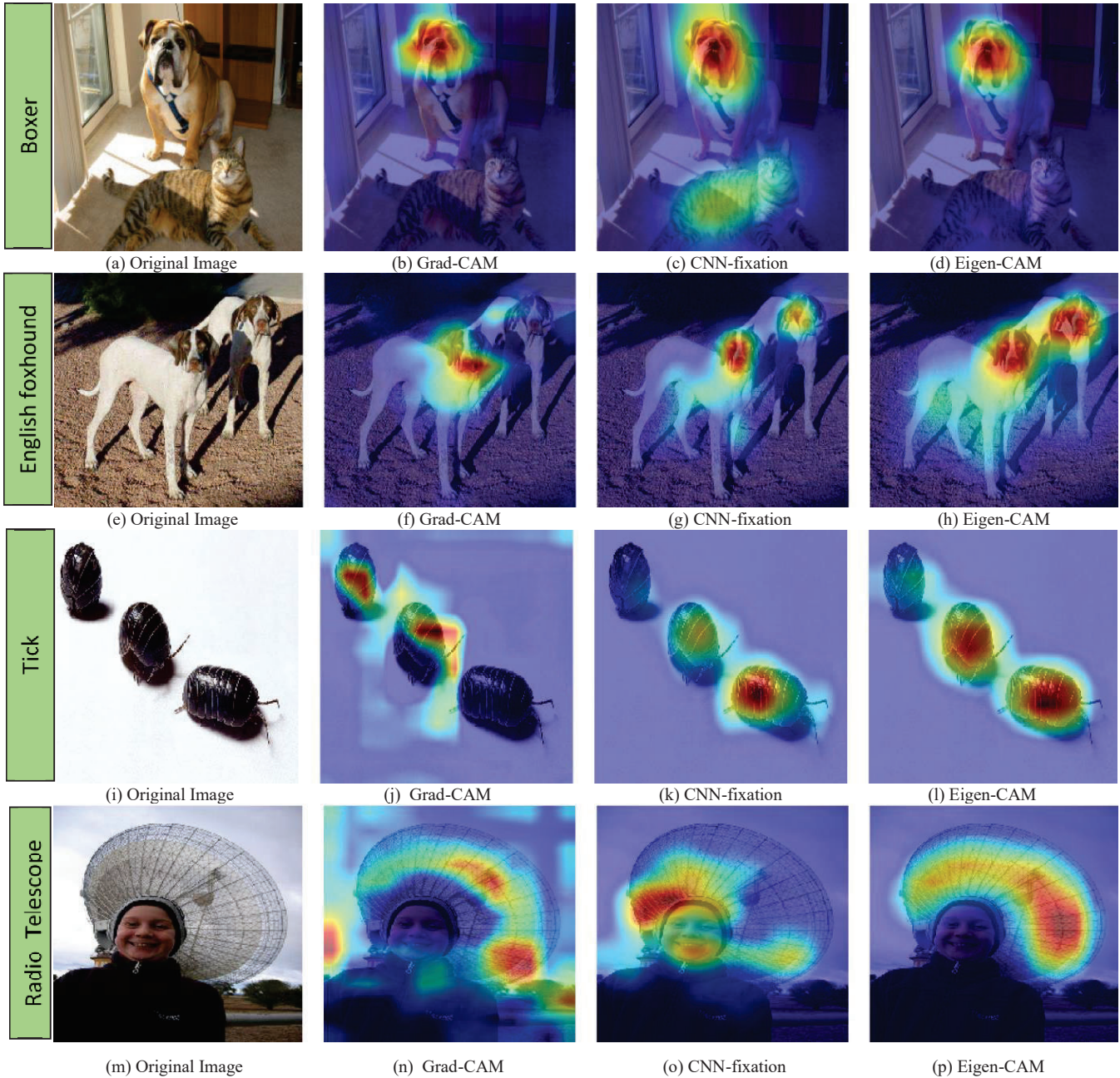

Fig. 2: CNN visualizations computed for a four sample images from ILSVRC validation set (a, e, I, m). Second column form the left images (b, f, j, n) show class activations computed using Grad-CAM. second column from the right images (c, g, k, o) represent activations produced using CNN Fixation. First column from the right images (d, h, l, p) show class activations computed using Eigen-CAM, all examples were classified correctly using VGG-16 and the green box represents the example label.

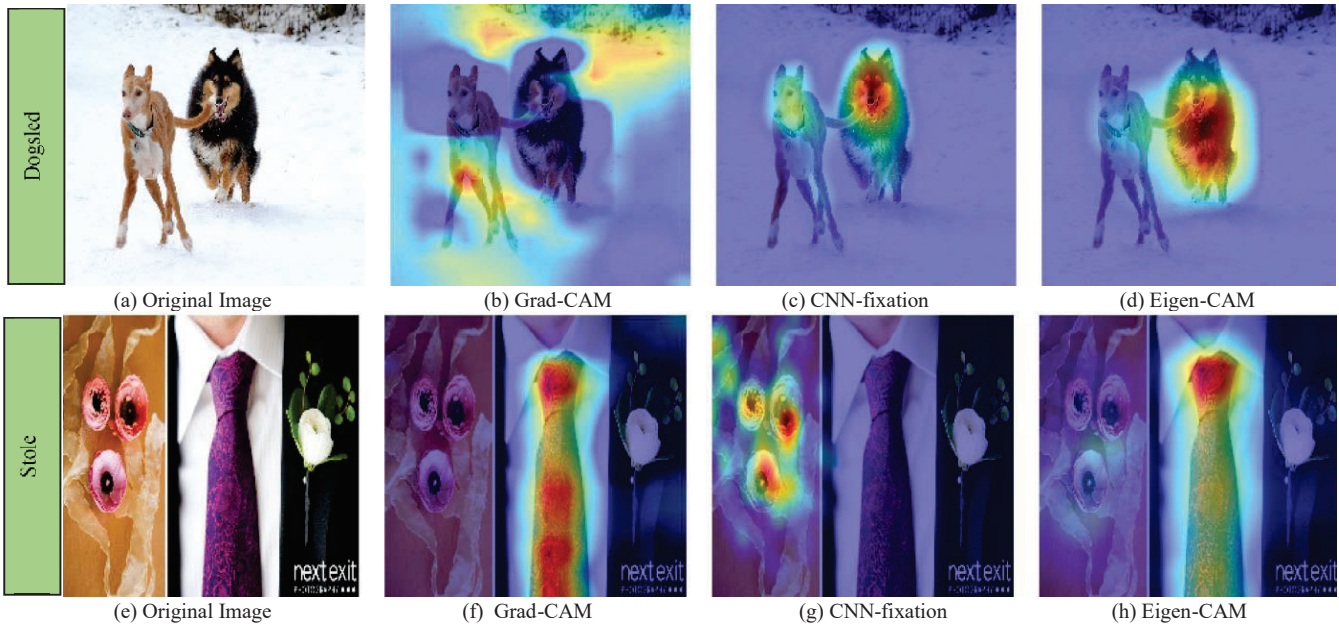

Fig. 3: CNN visualizations computed for a two misclassified example images from ILSVRC validation set (a, e) Original image. Second column form the left images (b, f) show class activations computed using Grad-CAM. second column from the right images (c, g) represent activations produced using CNN Fixation. First column from the right images (d, h) show class activations computed using Eigen-CAM, green box represents top 1 classification results using VGG-16

In the second example, top-3 classification results for the image in figure 3.e were ( Stole with probability 0.153, Kimono with probability 0.063 and Windsor tie with probability 0.055), and Eigen-CAM localizes features for Windsor tie as can be seen in figure 3.h.

In compliance with existing methods, we will demonstrate the effectiveness of localizing objects using Eigen-CAM across:

A. **Weakly-supervised localization**

In this section, we evaluated Eigen-CAM localization capability on the ILSVRC 2014 benchmark dataset (ImageNet) [28] in the context of image classification.

In weakly-supervised localization methods, CNNs utilize different techniques for object detection without training on bounding boxes, similar to other methods Eigen-CAM can localize objects, but rather than relying on classification labels to generate bounding boxes as in case of most methods, Eigen-CAM analyzed the output of the last convolutional layer to generate bounding boxes for the task of object localization.

For localization task, given an image and a model, in a forward pass, a class activation map is obtained from the first principle component of the combined weight matrix of the last convolution layer, scaled to (0 - 255) range, reshaped to a square size, And binarized based on different thresholds of (5-15)% of maximum level (255), different thresholds are caused by using different models, binarizing allow and facilitate producing a bounding box for the largest segment.

For the evaluation process, we used off-the-shelf pre-trained VGG-16 [29], AlexNe t[5], ResNet-101 [4], Inception-V1, aka GooleNet [25], and DenseNet-121 [19], all model are pre-trained on ILSVRC dataset, we evaluated these models on ILSVRC validation set of which include 50,000 images. We compare results obtained using the Eigen-CAM with all methods, as shown in Table 1. For all methods (except Grad-CAM), we used numbers reported in [22].

The results presented in Table 1 represent the error rate of Intersection over Union (IOU) metric (1 - Accuracy) for the top-1 recognition prediction, for all previous methods the computation of IOU metric was for the correct predictions only, unlike our method which is class independent and allows the computation of IOU for the entire validation set of ILSVRC dataset.

In the task of weakly-supervised localization, we achieved a 12% percent improvement using the AlexNet model, 7.5% percent improvement using the VGG-16 model, 11% percent improvement using GooLeNet model and 2.7 % percent improvement using DenseNet model.

B. **Adversarial Examples**

In this section, we will try to answer the question of which part of the CNN is mainly affected by adversarial examples. Adversarial examples can easily fool different classifiers while being imperceptible to humans, and adversaries deploy such vulnerability to impair different safety-critical environments that utilize CNNs.

To answer the question we started with, we have perturbed some examples from the ILSVRC dataset using the DeepFool method [30] for the VGG-16 Model, Figure 6. shows two Examples classified using the VGG-16 Model and perpetuated using the DeepFool method.

TABLE I
TOP-1 RECOGNITION PREDICTION ERROR RATES FOR THE WEAKLY SUPERVISED LOCALIZATION TASK OF DIFFERENT VISUALIZATION APPROACHES ON ILSVRC VALIDATION SET. BOLD FACE NUMBERS REPRESENT BEST RESULTS AMONG DIFFERENT MODELS

| Method | AlexNet | VGG-16 | DenseNet-121 | GoogLeNet | ResNet-101 |
|---|---|---|---|---|---|
| cMWP | 72.31 | 64.18 | 64.97 | 69.25 | 65.94 |
| Backprob | 65.17 | 61.12 | 67.49 | 61.31 | 57.97 |
| CAM | 67.19 | 57.20 | 55.37 | 60.09 | **48.34** |
| Grad-CAM | 71.16 | 56.51 | 75.29 | 74.26 | 64.84 |
| CNN Fixations | 65.70 | 55.22 | 56.72 | 57.53 | 54.31 |
| Eigen-CAM | **53.02** | **47.67** | **55.07** | **46.28** | 55.26 |

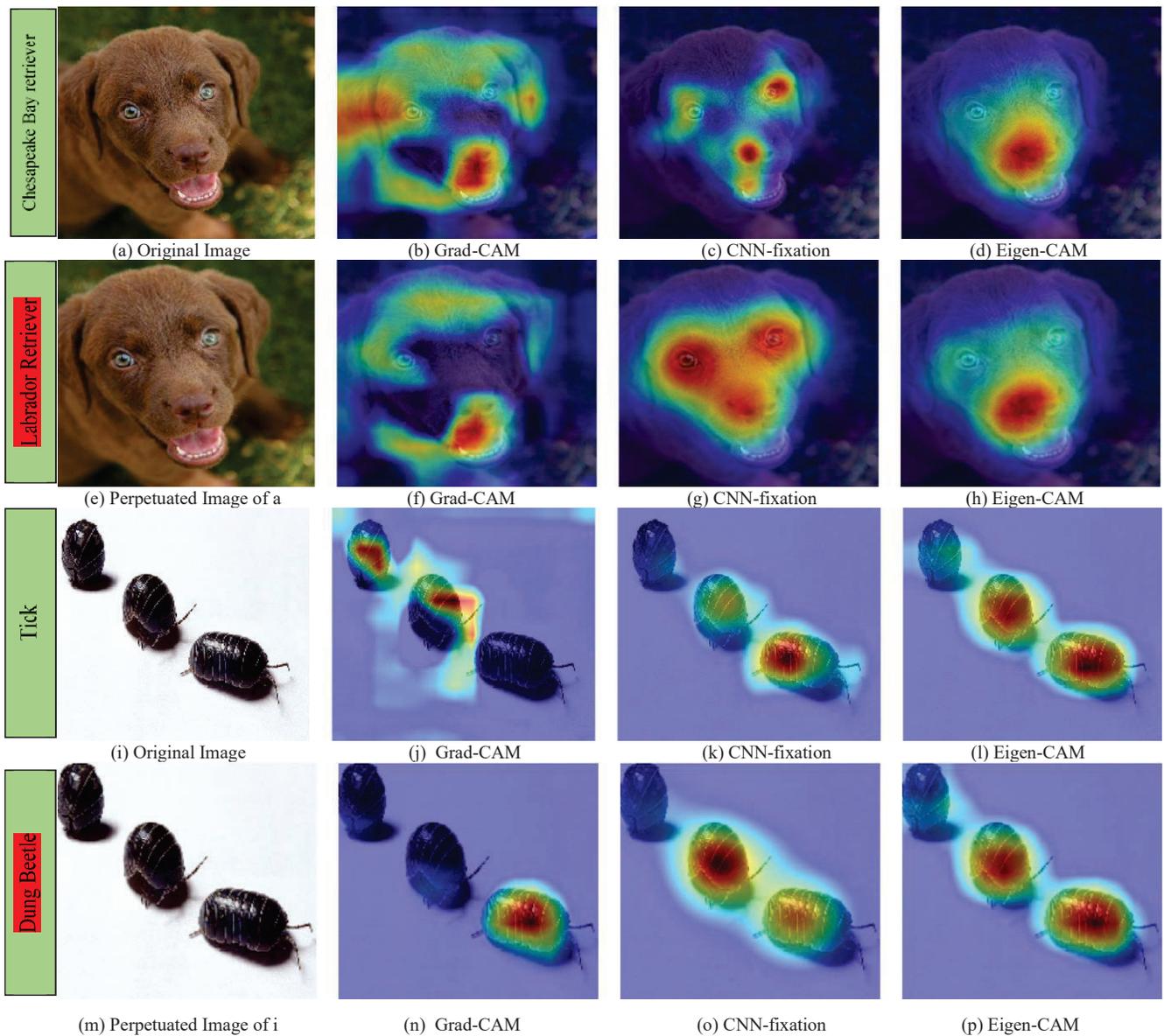

Fig. 4: CNN visualizations computed for two sample images from ILSVRC validation set (a, i) Original image. Image e and m are a perpetuated image of a and i Second column form the left images (b, f, j, n) shows class activations computed using Grad-CAM. second column from the right images (c, g, k, o) represent activations produced using CNN Fixation. First column from the right images (d, h, l, p) shows class activations computed using Eigen-CAM. green box represents the example label and red box represents the adversarial example classification result.

A closer examination of explanations produced on original and perpetuated images show that both the Grad-CAM (Figure 4.b vs. 4.f; 4.j vs. 4.n) and CNN fixation (Figure 4.c vs. 4.g; 4.k vs. 4.o) produce different activation maps. However, the Eigen-CAM produced almost identical visual explanation (see figure 4.d vs.4.h; 4.i vs. 4.p) and also correctly localize objects in the presence of adversarial noise. Since Eigen-CAM relies on the feature extraction network only and not on the classification network, we could infer that adversarial noise has more effect on classification network (dense layers), and that also explains the change in explanations for other methods.

## V. CONCLUSIONS

Deep CNN-based models are highly accurate yet difficult to explain. As a research community, we need easy to use (without any modifications of the model) method to gain perspective about the learned representations and their validations. We have presented the Eigen-CAM that provides visual explanation irrespective of the accuracy of the model or the presence of adversarial noise. We provide empirical evidence showing that Eigen-CAM is robust and reliable in producing consistent visual explanations and outperforms state-of-the-art methods. The easy to use and intuitive Eigen-CAM only needs the learned representations at the final convolution layer, making it independent of classification layers. Eigen-CAM can be used with any CNN-based DL models without any modification.